\documentclass{article}

\usepackage{arxiv}

\usepackage[utf8]{inputenc} % allow utf-8 input
\usepackage[T1]{fontenc}    % use 8-bit T1 fonts
\usepackage{hyperref}       % hyperlinks
\usepackage{url}            % simple URL typesetting
\usepackage{booktabs}       % professional-quality tables
\usepackage{amsfonts}       % blackboard math symbols
\usepackage{amsmath,amssymb}
\usepackage{nicefrac}       % compact symbols for 1/2, etc.
\usepackage{microtype}      % microtypography
\usepackage{textcomp}
\usepackage{enumitem}
\newcommand{\subscript}[2]{$#1 _ #2$}
\usepackage{xcolor}
\usepackage{tikz}
\usetikzlibrary{bayesnet}
\def\BibTeX{{\rm B\kern-.05em{\sc i\kern-.025em b}\kern-.08em
    T\kern-.1667em\lower.7ex\hbox{E}\kern-.125emX}}
    
\newcommand*{\prob}{\mathsf{P}}

\title{Bayesian Stress Testing of Models in a Classification Hierarchy}

\author{
  Bashar Awwad Shiekh Hasan\\
  Caspian Learning\\
  Newcastle Upon Tyne, UK\\
  \texttt{bashar.hasan@caspian.co.uk} \\
   \And
 Kate Kelly \\
  Caspian Learning\\
  Newcastle Upon Tyne, UK\\
  \texttt{kate.kelly@caspian.co.uk} \\
}

\begin{document}
\maketitle

\begin{abstract}
Building a machine learning solution in real-life applications often involves the decomposition of the problem into multiple models of various complexity. This has advantages in terms of overall performance, better interpretability of the outcomes, and easier model maintenance. In this work we propose a Bayesian framework to model the interaction amongst models in such a hierarchy. We show that the framework can facilitate stress testing of the overall solution, giving more confidence in its expected performance prior to active deployment. Finally, we test the proposed framework on a toy problem and financial fraud detection dataset to demonstrate how it can be applied for any machine learning based solution, regardless of the underlying modelling required.
\end{abstract}

% keywords can be removed

\section{Introduction}
\label{sec:intro}
Machine learning has seen in the last 5-10 years an explosion in its growth from a research centered area of computer science and mathematics to a driving force for innovation in every aspect of our lives \cite{gainza2019deciphering,mckinney2020international,chen2018machine}. This was driven mainly by the success of deep learning and the significant investment of big technology firms in open source machine learning research \cite{lecun2015deep,abadi2016tensorflow,paszke2019pytorch}.

Real life machine learning based solutions often require a number of models to work together to achieve the business goal of the product(s) \cite{sachan2018learning}. Such models can be trained independently or as part of an optimised training pipeline \cite{bunescu2008learning,zhang2016flash}. 

Breaking down the product into multiple models has several advantages: I) It allows for parallel model development with model designers focused on solving relatively small and well-defined problems. II) It provides more transparency on how the solution makes decisions by providing the end user with information about how the decision process has been broken down and how individual sub-decisions were made. III) It allows the business to customise and variate the product by replacing/adding models in the solution hierarchy to satisfy business or customer needs. However, this comes at a cost: I) Error propagation between models can be a serious issue with errors propagating quickly to undermine the performance of the overall solution \cite{finkel2006solving,polyzotis2018data}. II) In a fast-paced production environment it is not clear how to prioritise model improvement. III) An update of a model will have a cascading effect on the rest of the models and will hence require re-training of all dependent models.

Building machine learning based solutions goes further beyond the performance of the individual models in the solutions \cite{howard2012designing}. The authors in \cite{amershi2019software} discussed the different aspects of building machine learning solutions from a software engineering point of view and concluded some of the unique challenges in building such solutions especially in how to train and update the models. A rubric for machine learning production readiness is defined in \cite{46555} to help create reliable, production-level machine learning systems. In \cite{caspiantrust2019} we presented a risk based approach to the use of machine learning in anti money laundering which adds additional requirements to the solution including: model explainability, bias, confounds, etc.

Surprisingly, there has been very little work in the literature on how to model and understand the interaction amongst models within a solution. Some work focused on optimising the parameters of the pipeline \cite{zhang2016flash,marciniak2005beyond,sachan2018learning}. Others focused on the troubleshooting and diagnosis of faults and errors in a machine learning pipeline \cite{zhang2017diagnosing,bruckner2014ml,nushi2017human}. 

With the increased use of pre-trained models \cite{devlin2018bert,simon2016imagenet}, pipelining and putting models in a hierarchy has become the norm in most machine learning solutions. This makes the need to stress test classification hierarchy in machine learning solutions ever more pertinent to increase confidence in the solution and trust in the model predictions.

The authors in \cite{finkel2006solving} proposed an architecture for modelling a pipeline as a Bayesian network. Their work is written in the context of natural language processing (NLP), where the end-to-end performance of an NLP system is limited, and potentially degraded, by the performance of the models and the error propagation through the pipeline. In their proposed architecture, models within the pipeline are represented as variables in the network. By sampling the entire distribution over the labels at each stage of the pipeline, a probabilistic output can be produced, smoothing the output of the models and improving the end-to-end solution performance. This directly addresses the error propagation problem in NLP, but does not cover the rest of the challenges of building a class hierarchy/pipeline.

In this paper we present a general purpose framework for building a Bayesian network that can be used to perform stress testing on a set of models that are connected in a set hierarchy. The framework will help answer important questions not only to understand the behaviour of the built model but also to assist in its maintenance and ongoing development (see Section \ref{subsec:def}). The main contribution of this paper is not narrowly defined for a specific application area, e.g. NLP, but is general and abstract enough to apply to any machine learning solution regardless of the underlying modelling approach required.

After presenting the framework in Section \ref{sec:methods}, we present a toy problem to explain how the framework could be implemented in Section \ref{subsec:toy}. Section \ref{subsec:fraud} presents the results of applying the framework on a sample fraud detection dataset. Section \ref{sec:disscus} provides additional discussion on the framework and its potential use.

\section{Methods}
\label{sec:methods}
\subsection{Problem Definition}
\label{subsec:def}
Assume a machine learning based solution consisting of a hierarchy of $N$ models $M =\{m_1,...,m_N\}$. The models use raw $K$ data features $X=\{x_1,...,x_K\}$ and propagate sub decisions through the hierarchy towards a top model(s). The input features $X$ can be continuous or discrete, and each model takes as an input a subset of the features and the output of a number of models, $q \geq 0$. Each model produces either a class prediction or regression value.

The relationships amongst the models and the input features can be represented as a directed acyclic graph (DAG)  $G=(V,E)$, where $V=\{v_i: i \in [1,N+K]\}$ is a node in the graph which can be either a model or an input feature, and $E=\{e_{ij}: i\neq j \; and \; i,j \in [1,N+K]\}$ is the directed link from $v_i$ to $v_j$ which represents information flow in the edge direction.

We would like to be able to answer questions like:
\begin{enumerate}[label=(\subscript{Q}{{\arabic*}})]
    \item What is the impact of a change of the distribution of feature $x_k$ on the overall performance of the solution?
    \item What is the impact of updating a model $m_i$ on the overall performance of the solution?
    \item Which model in the hierarchy would be the most influential in improving the overall solution performance?
\end{enumerate}

\subsection{Model Definition}
\label{subsec:mdef}

A Bayesian framework is best fit to answer the questions addressed in this work. It leads to a generative model of the overall solution that allows for running simulations targeted at stress testing the solution. The hierarchy being represented as a DAG makes it easy to model the solution as a Bayesian belief network, which in turn allows the use of the full functionalities of the Bayesian framework - for example Bayes chain rule, prior distributions and sampling \cite{barber2012bayesian}.

Under the Bayesian framework, it is not enough for a model in the hierarchy to produce predictions of the most likely class, it must also produce a probabilistic output per class. A probabilistic output is intrinsic to Bayesian and probabilistic based classifiers \cite{barber2012bayesian}, but it can also be easily extended to most commonly used classification or regression models in the literature \cite{springenberg2016bayesian,williams2006gaussian,denker1991transforming, platt1999probabilistic}, and can even be extended to rule-based learning \cite{sulzmann2009study,dembczynski2008maximum}.

The framework gives us the flexibility of defining prior information about the models within the hierarchy. Those prior distributions help incorporate domain knowledge about the class distribution of the particular model within the solution. It is common to model a classifier output as a multinomial or categorical distribution. Dirichlet distribution is a conjugate prior to those distributions and hence is a good choice for a prior distribution for the models in our framework \cite{turkman2019computational,barber2012bayesian}, although there is no restriction on using only Dirichlet. Similarly, prior distributions can also be defined for the input features. We define $\Theta=\{\theta_i: i \in [1,N+K]\}$ as the set of the prior distributions for all the models and the features.

Following the Bayesian chain rule, the joint probability distribution of the models, features, and priors can then be defined as:
\begin{equation}
\label{eq:main}
    \prob(M,X,\Theta) = \prob(V,\Theta) = \prod_{i=1}^{N+K} \prob(v_i|A_i,\theta_i)\prob(\theta_i)
\end{equation}
where $v_i \in V$ and 
\begin{equation}
    A_i = \{v_j: \exists\; e_{ji} \in E\}
\end{equation}
is the set of the nodes that $v_i$ has dependency on. If $A_i = \emptyset$, then $ \prob(v_i|A_i,\theta_i) = \prob(v_i|\theta_i)$. If no prior is defined for $v_i$, then $\prob(v_i|A_i,\theta_i)=\prob(v_i|A_i)$ and $\prob(\theta_i)=1$.

The conditional probability distributions $\prob(v_i|A_i,\theta_i)$ are usually difficult to estimate. However, given that we are working with models that have already been trained, the conditional distribution is simply the prediction output of the model. Using the known distributions of the input features, $X$, the framework becomes a generative model that can propagate data through the models from the independent variables towards the top of the hierarchy.

\subsection{Inference}
Most classification models would have a relatively low number of classes. In which case, Equation \ref{eq:main} allows for an exact inference. However, in some applications and especially those involving natural language processing pipelines the number of possible values for the variables become too large that exact inference becomes intractable. For example, parsing, part of speech tagging, and named entity recognition have a number of possible classes that is exponential to the length of the input sentence. In which case approximate inference becomes a necessity \cite{finkel2006solving}. We recommend the use of a Markov chain Monte Carlo (MCMC) approach, e.g. Random Walk Metropolis \cite{roberts2004general}. 

Given a set of observable data $D$, the log posterior distribution can be defined as:
\begin{equation} \label{eq:posterior}
\begin{split}
\log(\prob(\Theta|V,D)) & \propto \log(\prob(D|V,\Theta)\prob(\Theta)) \\
 & \propto \sum_{i=1}^{N+K} \log(\prob(D|v_i,A_i,\theta_i)) + \sum_{i=1}^{N+K}\log(\theta_i)
\end{split}
\end{equation}

The only remaining piece to run MCMC is the ability to sample from the independent and joint distributions in the model. 
\subsection{Sampling}
Sampling from the independent variables, most likely the features $X$, is straightforward, as those are parameterized by the relevant $\theta_i$. The exception is when the space to sample from is exponentially large, e.g. the space of all parse trees, or the words in the English language. To overcome this problem \cite{finkel2006solving,johnson2007bayesian} has proposed sampling approaches for multiple examples of such cases. The details of those sampling methods are outside the scope of this paper, but interestingly the changes required to support such sampling are minimal. 

Sampling from the conditional distribution is straightforward. Given that those are the models $M$ within our solution, we only need to sample their input features and then run the model to obtain the output samples, which can be passed to higher level model(s) in the hierarchy. 

\subsection{Stress testing}
Once the model is built, stress testing the solution can be easily achieved. The model is general enough to support a wide range of potential questions important for the solution operation in the live environment.  In the following we answer the questions raised in Section \ref{subsec:def}.

\begin{enumerate}[label=(\subscript{Q}{{\arabic*}})]
    \item To measure the impact of the change of a feature $x_k$ on the overall solution, we sample from the changed distribution of $x_k$. Kullback–Leibler (KL) divergence \cite{barber2012bayesian} would then be used as a measure of how the output probability distribution of the solution changes accordingly. This is particularly helpful to gain understanding of how the solution will react as a result of expected changes in the input data (e.g. due to data drift) or sudden changes in the input data (e.g. due to a deployment of the solution in a new territory). We recommend sampling repeatedly from the input feature distributions and calculating the joint distribution Eq. \ref{eq:main} so a more robust measure is obtained.
    \item Similarly to (Q1), KL divergence can be used to measure the change in the distribution of the output model as a result of changing a model using sampled data. Additionally, a change in the area under the curve (AUC) for the model at the top of the hierarchy can be a very informative measure of impact of a given model change on the overall solution performance.
    \item In order to decide which model is the most influential in the hierarchy, we retrain each individual model, one by one, on randomly assigned labels. This forces the model to produce predictions at the experimental random level, changing the output distribution of that single model. The AUC of the final solution with this updated model is then obtained and compared to the previous AUC where that single model was not trained on random labels. The resulting reduction in the AUC is then a strong indication of the importance of the targeted model on the overall solution. When this process is repeated for each model in the hierarchy, we can then determine which model should be focused on next when improving the solution.
\end{enumerate}

\subsection{Implementation Considerations}
We implemented the framework using tensorflow probability \cite{dillon2017tensorflow}. In the current implementation we relied heavily on tensorflow models whether linear or neural networks. If using direct inference, then the code can easily be extended to none tensorflow models. However, when using MCMC, this is more challenging and requires further development. The code used to produce the results in this paper is available at https://github.com/Caspian-Ltd.

\section{Experiments and Results}
\label{sec:exp}
To demonstrate the proposed framework and how it can be used to stress test a machine learning based solution, we use a toy problem and a more realistic dataset for fraud detection in transactional banking data. 
\subsection{Toy Problem}
\label{subsec:toy}

\begin{figure}[ht]
%\vskip 0.2in
\begin{center}
\centerline{\includegraphics[width=10cm]{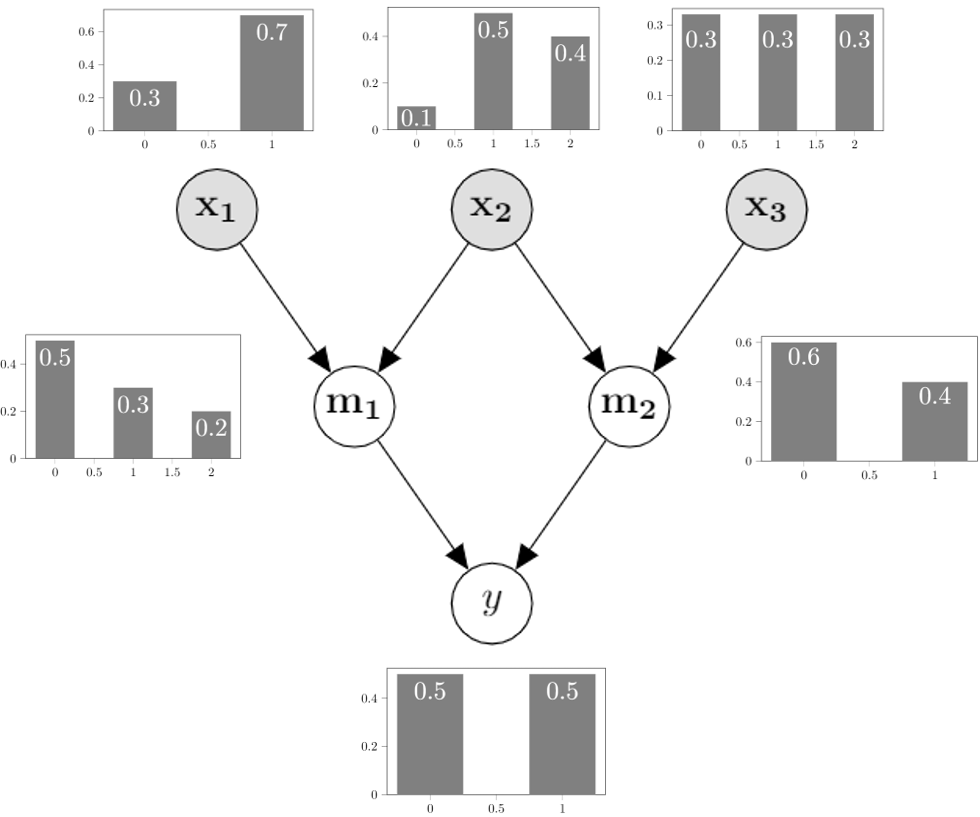}}
\caption{A toy example of a machine learning solution that uses a set of models in a hierarchical configuration. $\{x1,x2,x3\}$ are the raw features feeding into two classifiers $m_1$ and $m_2$, which in turn feed into the top model $y$. Each variable in the graph is represented with its categorical distribution.}
\label{fig:toy_example}
\end{center}
\vskip -0.2in
\end{figure}

The toy problem consists of three raw features $X= \{x_1,x_2,x_3\}$, with each feature having a categorical distribution, where $x_1$ has two categories, and $x_2$, $x_3$ both have three categories. The three models in the hierarchy are $M=\{m_1,m_2,y\}$, where $m_1$ is a classifier with 3-class output, and $m_2$, $y$ are both 2-class classifiers. The priors associated with those models are plotted in Fig \ref{fig:toy_example} along with the DAG.

 An equivalent DAG is built where the classifiers are replaced by categorical distributions, which are then used to generate sample data for training the models in $M$. In a real scenario this would be the available training data.

To establish a baseline, $m_1,m_2$ and $y$ are trained as linear classifiers and the joint probability is calculated following Eq. \ref{eq:main} if we are using exact inference, or using MCMC as described in Section \ref{sec:methods}.

\begin{figure}[ht]
%\vskip 0.2in
\begin{center}
\centerline{\includegraphics[width=14cm]{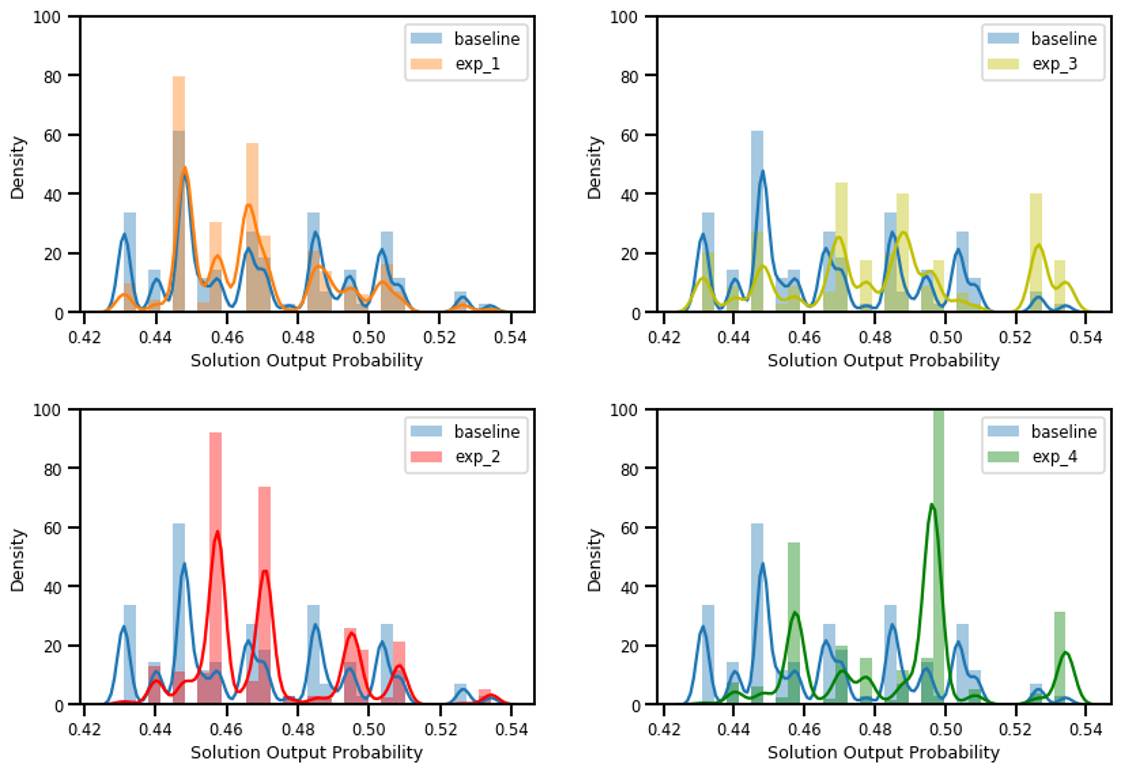}}
\caption{Normalized histograms of the solution output in the toy example as a result of the experiments 1-4. Each histogram is compared against the baseline distribution. In exp\textunderscore1 the distribution of $x_3$ is changed. exp\textunderscore2 changes the distributions of $x_1$ and $x_3$. exp\textunderscore3 modifies the distribution for $x_2$. exp\textunderscore4 changes all the input features.}
\label{fig:expirements}
\end{center}
\vskip 0in
\end{figure}

\textbf{\textit{Experiment 1:}}  Change the distribution of $x_3$ by setting the category probabilities to $[0.1,0.2,0.7]$. The simulation, i.e. sampling and joint probability estimation, is repeated 100 times. This results in KL-divergence of 0.21 indicating minor change in the output distribution as can be seen in Fig. \ref{fig:expirements} top left panel.

\textbf{\textit{Experiment 2:}}  In addition to change in Experiment 1, the distribution of $x_1$ is modified by setting the category probabilities to $[0.9,0.1]$. KL-divergence increases to 0.99 which is reflected in the shift from the baseline distribution, Fig. \ref{fig:expirements} bottom left panel.

\textbf{\textit{Experiment 3:}}  Sets the category probabilities for $x_2$ to $[0.6,0.3,0.1]$. KL-divergence is 0.35, Fig. \ref{fig:expirements} top right panel.

\textbf{\textit{Experiment 4:}} Combines the changes in Experiment 2 and 3. KL-divergence is 1.135 which can be seen as a much bigger change from the baseline distribution in Fig. \ref{fig:expirements} bottom right panel.

\textbf{\textit{Experiment 5:}} Here the linear classifier in $m_1$ is replaced by a two-layer feed forward Multi-Layer Perceptron.

\begin{figure}
%\vskip 0.2in
\begin{center}
\centerline{\includegraphics[width=14cm]{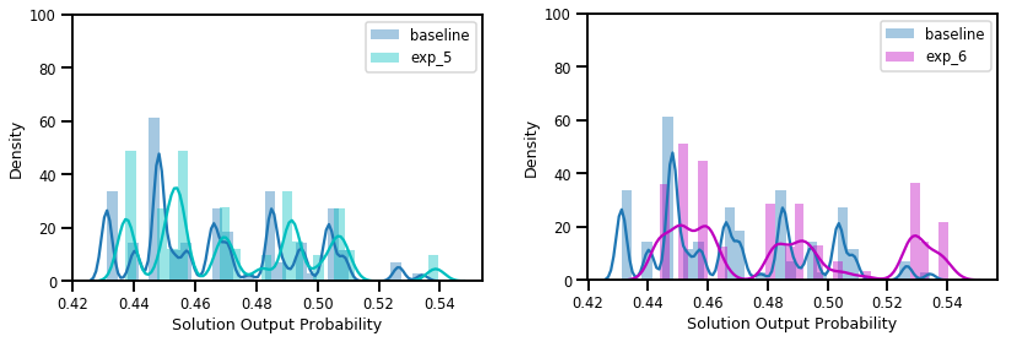}}
\caption{Normalized histograms of the solution output in the toy example as a result of the experiments 5-6 compared against the baseline distribution. In exp\textunderscore5 the linear classifier $m_1$ is replaced by a neural network, while the linear $m_2$ is swapped by a boosted tree in exp\textunderscore6.}
\label{fig:expirements_2}
\end{center}
\vskip -0.2in
\end{figure}

\textbf{\textit{Experiment 6:}} The linear classifier in $m_2$ is replaced by a boosted tree classifier. Figure \ref{fig:expirements_2} demonstrates the outcome of experiments 5 and 6. The MLP did not change the model performance compared to the linear model ($<$1\% difference in accuracy on testing set), but we can still see a change in the distribution of the final model output with KL-divergence of  $\sim 0.14$. The boosted tree does actually increase the accuracy of the $m_2$ by $\sim 10\%$. The AUC of the solution has increased by $4\%$ with KL-divergence $\sim 0.16$.

\begin{figure}
%\vskip 0.2in
\begin{center}
\centerline{\includegraphics[width=14cm]{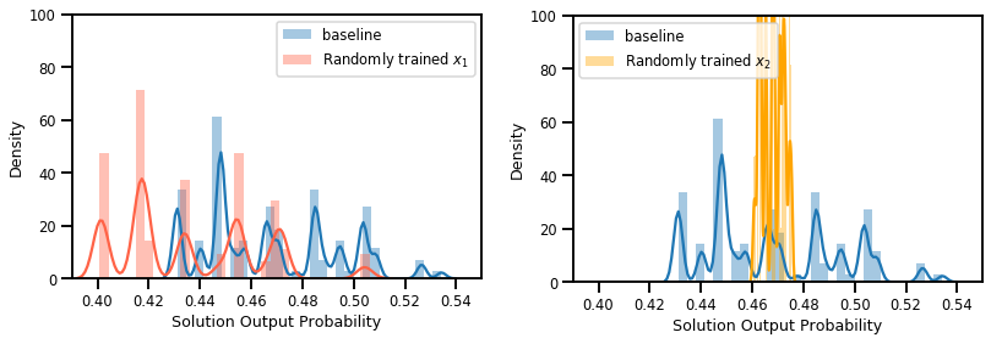}}
\caption{Normalized histograms of the solution output in the toy example as a result of experiment 7 compared against the baseline distribution. In the top panel the $m_1$ linear classifier is re-trained using randomly assigned labels. In the bottom panel the same is repeated for $m_2$.}
\label{fig:expirements_3}
\end{center}
\vskip -0.2in
\end{figure}

\textbf{\textit{Experiment 7:}} In order to test the impact of $m_1$, and $m_2$ on the overall performance of the solution, we separately re-train them using randomly assigned labels and measure the respective KL-divergence and change in AUC. Figure \ref{fig:expirements_3} shows the impact of this test. Replacing $m_1$ with an, essentially, random classifier does not seem to have made a significant change to the output of the solution: KL-divergence $\sim 0.11$ and a $\sim 2\%$ drop in AUC. On the other hand $m_2$ has a significant influence on the solution output, KL-divergence is very large ($\infty$) and the AUC showed a drop to chance level at $50\%$. This clearly suggests that $m_2$ is much more important for the overall solution performance compared to $m_1$, which is supported by the results of Experiment 6.

\subsection{Fraud Detection Example}
\label{subsec:fraud}

To test the framework on a more realistic dataset, we used the BankSim dataset \cite{lopez2014banksim}. BankSim is ``an agent-based simulator of bank payments based on a sample of aggregated transactional data provided by a bank in Spain''. The simulator is designed to generate synthetic data that can be used for the study of fraud detection. The dataset contains approximately six months worth of data and an average of two fraudulent transactions per day. In total there are 594,643 records with 587,443 normal payments and 7,200 fraudulent transactions.

There are 7 features in the data:
\begin{itemize}
    \item step: time step
    \item customer: the customer ID
    \item age: a numeric value 0-7 representing the age category of the customer
    \item gender: the gender of the customer
    \item merchant: the ID of the merchant in a given transaction
    \item category: the nature of business of the merchant
    \item amount: transaction amount
\end{itemize}

\begin{figure}[ht]

  \begin{center}
    \begin{tabular}{cc}
      % model_pca.tex
%
% Copyright (C) 2012 Jaakko Luttinen
%
% The MIT License
%
% See LICENSE file for more details.

% PCA model

%\beginpgfgraphicnamed{model-pca}
\begin{tikzpicture}

  \newlength\figureheight
  \newlength\figurewidth
    
  \setlength\figureheight{2in}
  \setlength\figurewidth{3in}

  % Define nodes
  \node[latent]                               (y) {y};
  
  \node[latent, above=of y, xshift=-2.4cm] (m_1) {$\mathbf{m_1}$};
  \node[obs, above=of y, xshift=-1.2cm] (x_1) {$\mathbf{x_1}$};
  \node[obs, above=of y, xshift=0cm]  (x_2) {$\mathbf{x_2}$};
  \node[obs, above=of y, xshift=1.2cm]  (x_3) {$\mathbf{x_3}$};
  \node[latent, above=of y, xshift=2.4cm]  (m_2) {$\mathbf{m_2}$};
  
  \node[obs, above=of m_1, xshift=0cm]  (x_4) {$\mathbf{x_4}$};
  \node[obs, above=of m_2, xshift=-2.4cm]  (x_5) {$\mathbf{x_5}$};
  \node[obs, above=of m_2, xshift=-1.2cm]  (x_6) {$\mathbf{x_6}$};
  \node[obs, above=of m_2, xshift=0cm]  (x_7) {$\mathbf{x_7}$};
  \node[obs, above=of m_2, xshift=1.2cm]  (x_8) {$\mathbf{x_8}$};
  \node[obs, above=of m_2, xshift=2.4cm]  (x_9) {$\mathbf{x_9}$};

  % Connect the nodes
  \edge {m_1, m_2, x_1, x_2, x_3} {y} ; %
  \edge {x_1, x_4} {m_1} ; %
  \edge {x_5, x_6, x_7, x_8, x_9} {m_2} ; %

\end{tikzpicture}
%\endpgfgraphicnamed

%%% Local Variables: 
%%% mode: tex-pdf
%%% TeX-master: "example"
%%% End: 
    \end{tabular}
  \end{center}
  \caption{Bayesian network represented as a DAG for Fraud Detection Example}
  \label{fig:fraud}
\end{figure}
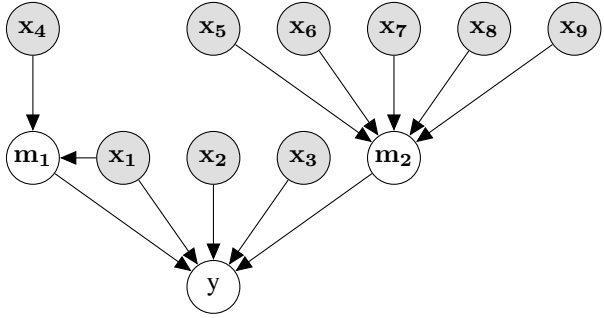

To solve the challenge in fraud detection in Banksim we have applied some background knowledge of the AML investigation process and developed a classification hierarchy that consists of three models: 
\begin{itemize}
    \item The Nature of Business Risk classifier ($m_1$): assesses the risk of transacting with the merchant and rates the transaction as \textit{low, medium, high} risk.
    \item Transaction Frequency classifier ($m_2$): assesses the frequency of the transaction with the merchant in the customer data. The classifier predicts whether the transaction is \textit{rare, infrequent, regular}.
    \item Decision classifier ($y$): uses the engineered features and $m_1$ and $m_2$ to predict whether the transaction is fraudulent or not.
\end{itemize}
To support the models several new features are engineered, in particular for the Transaction Frequency classifier. Table \ref{table:nodes} summarizes the features $X=\{x_1,...,x_9\}$, and Figure \ref{fig:fraud} demonstrates the DAG representing the classification hierarchy.

All models $M=\{m_1,m_2,y\}$ are trained as linear classifiers. To reduce the impact of imbalanced data (98\% of transactions are not fraudulent) the majority class is down-sampled when training the Decision classifier so we have an equal number of samples for both classes. To train $m_1$ and $m_2$ a randomly selected subset of the data is selected and manually labelled. Note that the choice of a linear classifier is not significant to the problem and may not necessarily be the best choice for the problem, however this experiment was not to find the best classifier, but was to display how the hierarchy of models can be utilised in a Bayesian framework.
% Fraud network graph

\begin{table}[ht]
  \caption{Description of the node types in the Bayesian network of the fraud detection example}
  \label{table:nodes}
  \begin{center}
    \begin{tabular}{|p{0.75cm}|p{4.75cm}|p{2.5cm}|}
      \hline
      Node & Details & Distributions 
      \\
      \hline
      $x_1$ &
      Normalized Transaction Amount & Truncated Normal
      \\
      $x_2$ &
      Customer Age & Categorical
      \\
      $x_3$ &
      Customer Gender & Categorical
      \\
      $x_4$ &
      Category & OneHot Categorical
      \\
      $x_5$ &
      Average Time Between Transactions with a merchant & Gamma
      \\
      $x_6$ &
      Standard Deviation of Time Between Transactions with a merchant & Gamma
      \\
      $x_7$ &
      Average Transaction Amount with a merchant & Gamma
      \\
      $x_8$ &
      Standard Deviation of Transaction Amount with a merchant & Gamma
      \\
      $x_9$ &
      Ratio of the number of transactions with a merchant to the number of transactions the customer has & Truncated Normal
      \\
     $m_1$ &
      Nature of Business Risk Classifier & OneHot Categorical
      \\
      $m_2$ &
      Transaction Frequency Classifier & OneHot Categorical
      \\
      $y$ &
      Decision Classifier & OneHot Categorical
      \\
      \hline
    \end{tabular}
  \end{center}
\end{table}

Similar to the toy example, we ran a number of experiments to demonstrate the potential use cases of the framework. We will focus on the distribution of the ``Fraud'' class as, from our experience in the domain, this is a key factor for the investigating banks. High number of predicted fraudulent transactions will lead to a large false positive rate which in turn leads to significant operational overhead to the bank. On the other hand low predictions might lead to missing true fraudulent activity. Naturally a high f1 score is the preference but if a compromise to be made then a higher recall is usually preferred to high precision. 

\begin{figure}
%\vskip 0.2in
\begin{center}
\centerline{\includegraphics[width=12cm]{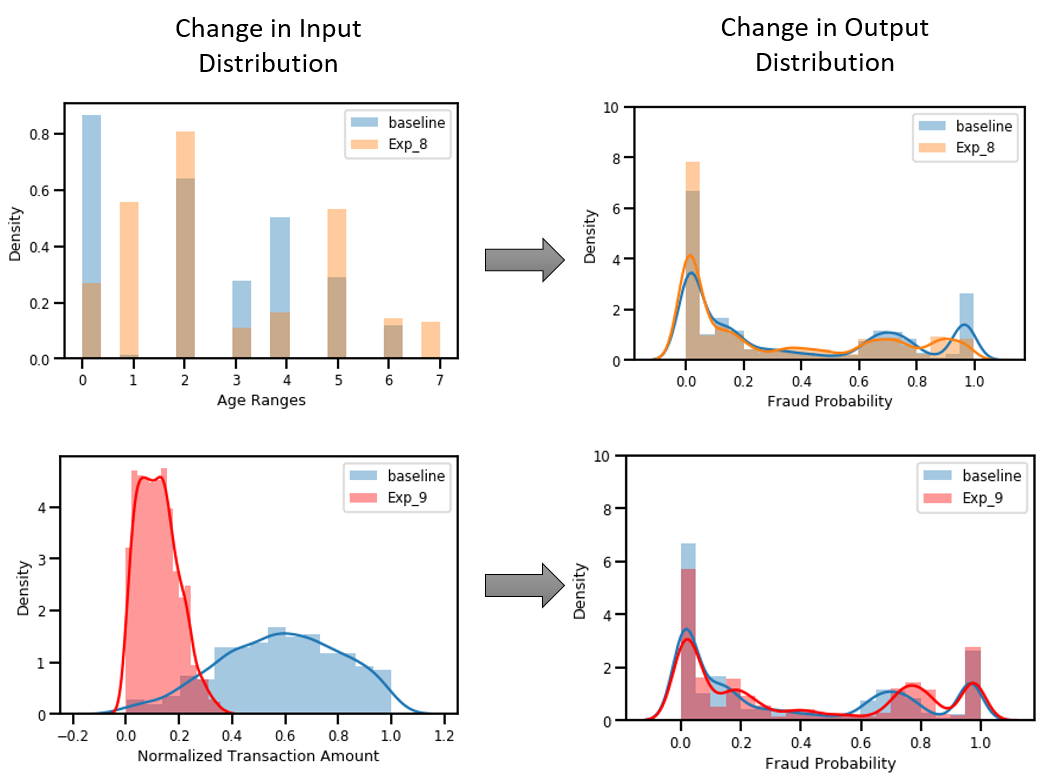}}
\caption{Normalized histograms of demonstrating the change in the distributions of input features Age ($x_2$) and Transaction Amount($x_1$), and the resulted change in the output distribution.}
\label{fig:expirements_6}
\end{center}
\vskip -0.2in
\end{figure}

\textbf{\textit{Experiment 8:}} The distribution of customer's age ($x_2$) is modified. We performed 100 repetition of the sampling (5000 samples) and joint probability estimation (Eq. \ref{eq:main}). The difference in $x_2$ distribution and the resulted distribution of the ``Fraud'' class is shown in Fig. \ref{fig:expirements_6} (upper panel). KL-divergence is 0.024 between the baseline and the resulted ``Fraud'' probability distribution.

\textbf{\textit{Experiment 9:}} The distribution of the normalized transaction amount ($x_1$) is modified to produce much smaller values. As in Experiment 8, we performed 100 repetition of the sampling and joint probability estimation. The change in $x_1$ distribution and the output ``Fraud'' probability is shown in Fig. \ref{fig:expirements_6} (lower panel). KL-divergence is 0.007 between the baseline and the resulted ``Fraud'' probability distribution indicating little change in the predicted risk.

\begin{figure}
%\vskip 0.2in
\begin{center}
\centerline{\includegraphics[width=10cm]{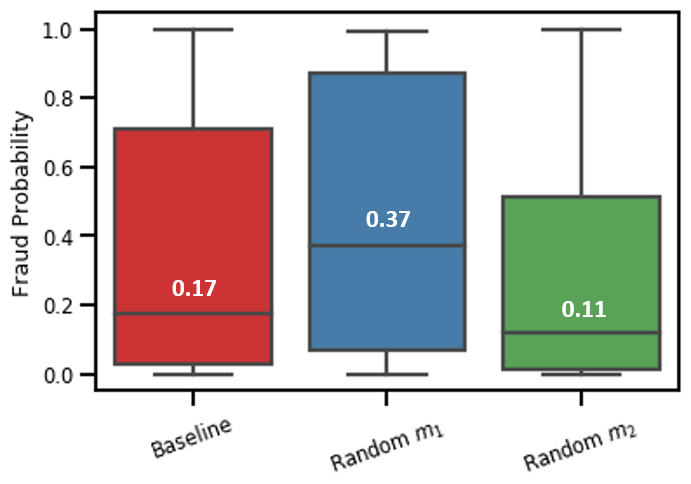}}
\caption{Box plot of the fraud probability as a result of training the nature of business classifier $m_1$ on randomly labelled data and the training of the frequency classifier $m_2$ on random labels. The median value of each case is overlaid on the box plots.}
\label{fig:expirements_7}
\end{center}
\vskip -0.2in
\end{figure}

\textbf{\textit{Experiment 10:}} Here we are trying to understand the relevant importance of the Nature of Business Risk classifier $m_1$ and the Transaction Frequency classifier $m_2$ to the overall solution performance. Both models are retrained with randomly assigned labels to the training data in order to produce experimental chance level results. The new models are then, separately, replaced by the baseline models and a 100 repetition of the simulations are ran.  Fig. \ref{fig:expirements_7} uses box plots to show the resulted probability of ``Fraud'' in the three scenarios: baseline, $m_1$ with randomly assigned labels, and $m_2$ with randomly assigned labels. It is clear from the figure that an unreliable Nature of Business Risk classifier leads to an increase in the predicted fraudulent transactions as the median of the fraud probability significantly shifts upwards. This, however, results in a 9\% reduction in recall indicating that the overall performance of the solution has dropped significantly. On the other hand a random Transaction Frequency classifier results in a 6 \% drop in the median of the ``Fraud'' probability with a 9\% reduction in recall. This implies that in this case the solution is significantly underestimating risks.

\begin{figure}
%\vskip 0.2in
\begin{center}
\centerline{\includegraphics[width=14cm]{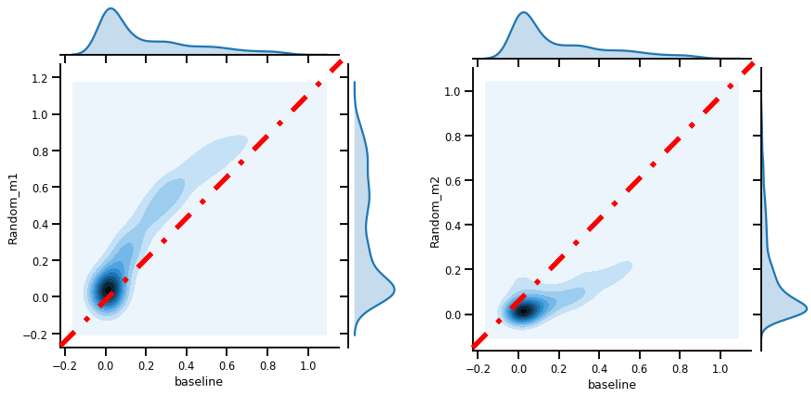}}
\caption{The bi-variate distribution of the probability of fraud of baseline solution against the models in Experiment 10. Random $m_1$ is the result of replacing $m_1$ with a classifier trained on randomly permuted labels of nature of business risk. Random $m_2$ is the result of replacing $m_2$ with a classifier trained on randomly permuted labels for frequency. }
\label{fig:expirements_8}
\end{center}
% \vskip -0.2in
\end{figure}

Figure \ref{fig:expirements_8} demonstrates further the impact of changes in Experiment 10 on the output of the solution. It is clear from the top panel that a poorly performing Nature of Business Risk classifier leads to an increase in reported fraud (mostly incorrectly), while a poorly performing Transaction Frequency model has the opposite impact. This demonstrates the power of the proposed framework as it allows a decision maker to understand the behaviour of their fraud detection system. It also empowers the risk managers to define the types of risks that might occur as a result of adopting a machine learning solution and then enforce the required measures and processes to mitigate those risks.

\section{Discussion}
\label{sec:disscus}
In this work we proposed a general purpose framework for stress testing machine learning based solutions. The framework takes a Bayesian approach to modelling a hierarchy of classifiers as a Bayesian network with directed links between nodes indicating probabilistic dependencies. This allows the use of the full Bayesian framework for parameter estimation and inference. After defining the model, we demonstrated how the framework can be used by applying it on a simplified toy example, and then on a realistic problem: fraud detection in banking systems.

The main purpose of the framework is to increase the trust in the deployed machine learning solution before going live and during the ongoing maintenance. In real-life applications, labelled data is very difficult to obtain and even if labels are available they usually come after a lengthy labelling process. Subject matter experts are usually used to evaluate periodically and retrospectively the outcome of the machine learning solution. This leaves a gap of uncertainty regarding the performance of the solution especially in highly risky environments, for example, in fraud detection and anti-money laundering. 

The proposed framework provides decision makers with directional evidence of how their deployed solution will most likely behave in extreme edge cases, or as a result of re-deploying the system in a different region with different data distributions. This is extremely valuable information for risk stewards, operation managers, and data science teams, as it provides insight to the solution performance in different scenarios whilst also minimising the risk involved.

In the examples provided here we assumed, without loss of generality, that the input features $X$ are independent, hence no links between them in Fig. \ref{fig:toy_example} and Fig. \ref{fig:fraud}. During the engineering of the machine learning solution, features are preferred to be independent. However this is not always possible, and in some situations dependencies are not very clear. The framework presented here is not impacted by any dependencies among the features as they are treated as any other node in the Bayesian network and the same Bayesian chain rule will apply. For highly complex problems, it might be more efficient to simulate the features independently and provide the simulation data to the framework here. This can be for example using agent-based simulations similar to \cite{lopez2014banksim}.

The built Bayesian model can be used as a basis for a causal model \cite{pearl2018book}. This will help provide further stress testing of the solution and potentially more robust interrogation of the models through interventions. In the future, this is the direction where we want to drive and build upon this work.

\bibliographystyle{unsrt}  
\bibliography{references}

\begin{thebibliography}{10}

\bibitem{gainza2019deciphering}
P~Gainza, F~Sverrisson, F~Monti, E~Rodol{\`a}, D~Boscaini, MM~Bronstein, and
  BE~Correia.
\newblock Deciphering interaction fingerprints from protein molecular surfaces
  using geometric deep learning.
\newblock {\em Nature Methods}, pages 1--9, 2019.

\bibitem{mckinney2020international}
Scott~Mayer McKinney, Marcin Sieniek, Varun Godbole, Jonathan Godwin, Natasha
  Antropova, Hutan Ashrafian, Trevor Back, Mary Chesus, Greg~C Corrado, Ara
  Darzi, et~al.
\newblock International evaluation of an ai system for breast cancer screening.
\newblock {\em Nature}, 577(7788):89--94, 2020.

\bibitem{chen2018machine}
Zhiyuan Chen, Ee~Na Teoh, Amril Nazir, Ettikan~Kandasamy Karuppiah, Kim~Sim
  Lam, et~al.
\newblock Machine learning techniques for anti-money laundering (aml) solutions
  in suspicious transaction detection: a review.
\newblock {\em Knowledge and Information Systems}, 57(2):245--285, 2018.

\bibitem{lecun2015deep}
Yann LeCun, Yoshua Bengio, and Geoffrey Hinton.
\newblock Deep learning.
\newblock {\em nature}, 521(7553):436, 2015.

\bibitem{abadi2016tensorflow}
Mart{'\i}n Abadi, Paul Barham, Jianmin Chen, Zhifeng Chen, Andy Davis, Jeffrey
  Dean, Matthieu Devin, Sanjay Ghemawat, Geoffrey Irving, Michael Isard, et~al.
\newblock Tensorflow: A system for large-scale machine learning.
\newblock In {\em 12th Symposium on Operating Systems Design and Implementation
  (OSDI 16)}, pages 265--283, 2016.

\bibitem{paszke2019pytorch}
Adam Paszke, Sam Gross, Francisco Massa, Adam Lerer, James Bradbury, Gregory
  Chanan, Trevor Killeen, Zeming Lin, Natalia Gimelshein, Luca Antiga, et~al.
\newblock Pytorch: An imperative style, high-performance deep learning library.
\newblock In {\em Advances in Neural Information Processing Systems}, pages
  8024--8035, 2019.

\bibitem{sachan2018learning}
Mrinmaya Sachan, Kumar~Avinava Dubey, Tom~M Mitchell, Dan Roth, and Eric~P
  Xing.
\newblock Learning pipelines with limited data and domain knowledge: a study in
  parsing physics problems.
\newblock In {\em Advances in Neural Information Processing Systems}, pages
  140--151, 2018.

\bibitem{bunescu2008learning}
Razvan~C Bunescu.
\newblock Learning with probabilistic features for improved pipeline models.
\newblock In {\em Proceedings of the Conference on Empirical Methods in Natural
  Language Processing}, pages 670--679. Association for Computational
  Linguistics, 2008.

\bibitem{zhang2016flash}
Yuyu Zhang, Mohammad~Taha Bahadori, Hang Su, and Jimeng Sun.
\newblock Flash: fast bayesian optimization for data analytic pipelines.
\newblock In {\em Proceedings of the 22nd ACM SIGKDD International Conference
  on Knowledge Discovery and Data Mining}, pages 2065--2074. ACM, 2016.

\bibitem{finkel2006solving}
Jenny~Rose Finkel, Christopher~D Manning, and Andrew~Y Ng.
\newblock Solving the problem of cascading errors: Approximate bayesian
  inference for linguistic annotation pipelines.
\newblock In {\em Proceedings of the 2006 Conference on Empirical Methods in
  Natural Language Processing}, pages 618--626. Association for Computational
  Linguistics, 2006.

\bibitem{polyzotis2018data}
Neoklis Polyzotis, Sudip Roy, Steven~Euijong Whang, and Martin Zinkevich.
\newblock Data lifecycle challenges in production machine learning: a survey.
\newblock {\em ACM SIGMOD Record}, 47(2):17--28, 2018.

\bibitem{howard2012designing}
Jeremy Howard, Margit Zwemer, and Mike Loukides.
\newblock {\em Designing great data products}.
\newblock " O'Reilly Media, Inc.", 2012.

\bibitem{amershi2019software}
Saleema Amershi, Andrew Begel, Christian Bird, Robert DeLine, Harald Gall, Ece
  Kamar, Nachiappan Nagappan, Besmira Nushi, and Thomas Zimmermann.
\newblock Software engineering for machine learning: a case study.
\newblock In {\em Proceedings of the 41st International Conference on Software
  Engineering: Software Engineering in Practice}, pages 291--300. IEEE Press,
  2019.

\bibitem{46555}
Eric Breck, Shanqing Cai, Eric Nielsen, Michael Salib, and D.~Sculley.
\newblock The ml test score: A rubric for ml production readiness and technical
  debt reduction.
\newblock In {\em Proceedings of IEEE Big Data}, 2017.

\bibitem{caspiantrust2019}
B.~Awwad Shiekh~Hasan J.~Faith and A.~Enshaie.
\newblock Trusting machine learning in anti-money laundering: A risk-based
  approach.
\newblock Technical report, Caspian Learning, Newcastle Upon Tyne, UK, 2019.

\bibitem{marciniak2005beyond}
Tomasz Marciniak and Michael Strube.
\newblock Beyond the pipeline: Discrete optimization in nlp.
\newblock In {\em Proceedings of the Ninth Conference on Computational Natural
  Language Learning (CoNLL-2005)}, pages 136--143, 2005.

\bibitem{zhang2017diagnosing}
Zhao Zhang, Evan~R Sparks, and Michael~J Franklin.
\newblock Diagnosing machine learning pipelines with fine-grained lineage.
\newblock In {\em Proceedings of the 26th International Symposium on
  High-Performance Parallel and Distributed Computing}, pages 143--153. ACM,
  2017.

\bibitem{bruckner2014ml}
Daniel Bruckner.
\newblock Ml-o-scope: a diagnostic visualization system for deep machine
  learning pipelines.
\newblock Technical report, CALIFORNIA UNIV BERKELEY DEPT OF ELECTRICAL
  ENGINEERING AND COMPUTER SCIENCES, 2014.

\bibitem{nushi2017human}
Besmira Nushi, Ece Kamar, Eric Horvitz, and Donald Kossmann.
\newblock On human intellect and machine failures: Troubleshooting integrative
  machine learning systems.
\newblock In {\em Thirty-First AAAI Conference on Artificial Intelligence},
  2017.

\bibitem{devlin2018bert}
Jacob Devlin, Ming-Wei Chang, Kenton Lee, and Kristina Toutanova.
\newblock Bert: Pre-training of deep bidirectional transformers for language
  understanding.
\newblock {\em arXiv preprint arXiv:1810.04805}, 2018.

\bibitem{simon2016imagenet}
Marcel Simon, Erik Rodner, and Joachim Denzler.
\newblock Imagenet pre-trained models with batch normalization.
\newblock {\em arXiv preprint arXiv:1612.01452}, 2016.

\bibitem{barber2012bayesian}
David Barber.
\newblock {\em Bayesian reasoning and machine learning}.
\newblock Cambridge University Press, 2012.

\bibitem{springenberg2016bayesian}
Jost~Tobias Springenberg, Aaron Klein, Stefan Falkner, and Frank Hutter.
\newblock Bayesian optimization with robust bayesian neural networks.
\newblock In {\em Advances in Neural Information Processing Systems}, pages
  4134--4142, 2016.

\bibitem{williams2006gaussian}
Christopher~KI Williams and Carl~Edward Rasmussen.
\newblock {\em Gaussian processes for machine learning}, volume~2.
\newblock MIT press Cambridge, MA, 2006.

\bibitem{denker1991transforming}
John~S Denker and Yann Lecun.
\newblock Transforming neural-net output levels to probability distributions.
\newblock In {\em Advances in neural information processing systems}, pages
  853--859, 1991.

\bibitem{platt1999probabilistic}
John Platt et~al.
\newblock Probabilistic outputs for support vector machines and comparisons to
  regularized likelihood methods.
\newblock {\em Advances in large margin classifiers}, 10(3):61--74, 1999.

\bibitem{sulzmann2009study}
Jan-Nikolas Sulzmann and Johannes F{\"u}rnkranz.
\newblock A study of probability estimation techniques for rule learning.
\newblock {\em FROM LOCAL PATTERNS TO GLOBAL MODELS}, page 123, 2009.

\bibitem{dembczynski2008maximum}
Krzysztof Dembczy{\'n}ski, Wojciech Kot{\l}owski, and Roman S{\l}owi{\'n}ski.
\newblock Maximum likelihood rule ensembles.
\newblock In {\em Proceedings of the 25th international conference on Machine
  learning}, pages 224--231. ACM, 2008.

\bibitem{turkman2019computational}
M~Ant{\'o}nia~Amaral Turkman, Carlos~Daniel Paulino, and Peter M{\"u}ller.
\newblock {\em Computational Bayesian Statistics: An Introduction}, volume~11.
\newblock Cambridge University Press, 2019.

\bibitem{roberts2004general}
Gareth~O Roberts, Jeffrey~S Rosenthal, et~al.
\newblock General state space markov chains and mcmc algorithms.
\newblock {\em Probability surveys}, 1:20--71, 2004.

\bibitem{johnson2007bayesian}
Mark Johnson, Thomas Griffiths, and Sharon Goldwater.
\newblock Bayesian inference for pcfgs via markov chain monte carlo.
\newblock In {\em Human Language Technologies 2007: The Conference of the North
  American Chapter of the Association for Computational Linguistics;
  Proceedings of the Main Conference}, pages 139--146, 2007.

\bibitem{dillon2017tensorflow}
Joshua~V Dillon, Ian Langmore, Dustin Tran, Eugene Brevdo, Srinivas Vasudevan,
  Dave Moore, Brian Patton, Alex Alemi, Matt Hoffman, and Rif~A Saurous.
\newblock Tensorflow distributions.
\newblock {\em arXiv preprint arXiv:1711.10604}, 2017.

\bibitem{lopez2014banksim}
Edgar~Alonso Lopez-Rojas and Stefan Axelsson.
\newblock Banksim: A bank payments simulator for fraud detection research.
\newblock In {\em 26th European Modeling and Simulation Symposium, EMSS}, pages
  144--152, 2014.

\bibitem{pearl2018book}
Judea Pearl and Dana Mackenzie.
\newblock {\em The book of why: the new science of cause and effect}.
\newblock Basic Books, 2018.

\end{thebibliography}
\end{document}